# Nonlinear Receding-Horizon Control of Rigid Link Robot Manipulators

**Hedjar, R. & Boucher, P.**
Supélec, Plateau de Moulon F91192 Gif Sur Yvette Cedex, France
hedjar@hotmail.com, patrick.boucher@supelec.fr

*Abstract: The approximate nonlinear receding-horizon control law is used to treat the trajectory tracking control problem of rigid link robot manipulators. The derived nonlinear predictive law uses a quadratic performance index of the predicted tracking error and the predicted control effort. A key feature of this control law is that, for their implementation, there is no need to perform an online optimization, and asymptotic tracking of smooth reference trajectories is guaranteed. It is shown that this controller achieves the positions tracking objectives via link position measurements. The stability convergence of the output tracking error to the origin is proved. To enhance the robustness of the closed loop system with respect to payload uncertainties and viscous friction, an integral action is introduced in the loop. A nonlinear observer is used to estimate velocity. Simulation results for a two-link rigid robot are performed to validate the performance of the proposed controller.*

## 1. Introduction

During recent years much emphasis has been placed on flexible manufacturing processes where the most important factors are quality, costs and time. Both fast motion in unconstrained space and mechanical interaction with the environment are required in most manufacturing systems. Industrial robots are often used to meet this demand and to perform various tasks such as material assembling, painting or welding. To accomplish these tasks efficiently and accurately, several control approaches have been proposed in the literature. Among these, a simple PD-control scheme that achieves satisfactory performance (Spong M. W. & Vidyasagar M. 1989) in the absence of gravity. However, robot manipulator is highly nonlinear system with coupling between joints and the gravity effects. The computed torque control or feedback linearization control has been also used to achieve best tracking performance. The implantation of the computed torque controller requires exact knowledge of the robot dynamics. Unfortunately, model uncertainties are frequently encountered in robotics due to unknown or changing payload and friction. These model uncertainties may decrease significantly the performance of this method in terms of tracking accuracy. Therefore, to achieve acceptable performance, even when all kinds of uncertainties are encountered, numerous robust control algorithms have been used like the variable structure approach (Slotine J. J. E. & Sastry S. S. 1983), robust adaptive approach (Ortega R. & Spong M. W. 1989), (Lee K. W. & Khalil H. K. 1997), (Canudas C. W. & Fixot N. 1992), (Spong M. W. 1992) and nonlinear $H_\infty$ approach (Chen B. S. *et al* 1994). A first survey of early results in robust control has been compiled in (Abdullah C. *et al*; 1991) and the second survey of recent results has been given in (Sage H. G. *et al*, 1999). Finally the robotic applications require effective control laws that achieve accurate tracking of fast motion despite the variations of inertia and gravitational load of the manipulator during operation.

Model predictive control of linear systems has received considerable attention in the last decade due to its robustness with respect to model uncertainties. However, many systems are inherently nonlinear. Since linear models are often inadequate to describe accurately the process dynamics, then nonlinear models should be used. Much effort has been made to extend linear predictive control to nonlinear systems (Michalska H & Mayne D. Q. 1993). The disadvantages of the proposed approach is the heavy online computation burden that causes two important problems in implementation of the nonlinear predictive control. One is the computation delay that cannot be ignored and the other is the global solution that cannot be guaranteed in each optimization problem. The application of these nonlinear control laws to nonlinear systems characterized by fast dynamics (such robotics ) sound like unusual proposal.  To overcome the computation burden, several nonlinear predictive laws have been developed in (Ping L. 1995), (Singh S. M. 1995), (Souroukh M. & Kravaris C. 1996), (Chen W. H. *et al*, 2003), where the one step ahead predictive output error is obtained by expanding the output signal and the is used to derive the offline control laws.

In this paper, the nonlinear receding-horizon controller proposed in (Ping L. 1998) is applied to robot manipulator to achieve position angular tracking objectives. To derive the control law, the predictive tracking error and the predicted control effort are minimized over a fixed time horizon. This approximate nonlinear controller is given in a closed form and thus no online optimization is required. Moreover, to increase the robustness of the control algorithm with regard to model uncertainties, we propose to introduce an integral action in the loop. The well-known Lyapunov based theory is used to show the asymptotic stability of the closed loop system in matched or mismatched case.

The major drawback of the proposed schemes is the requirement of measurement of motor speed. Speed measurements increase cost and impose constraints on the achievable bandwidth. Thus, to overcome this problem a nonlinear observer is used to estimate position and velocity angular of robot manipulator.

The outline of this paper proceeds as follows. In the next section, a dynamic model of robot manipulator is presented. In section 3, the approximate receding-horizon control scheme is developed to allow position angular tracking of a desired references trajectory. Stability analysis and robustness are treated in section 4. The high gain observer used to estimate unmeasured output elements (velocity angular) is presented in section 5. Simulation results are given in section 6. In the last, we conclude with some remarks.

## 2. Dynamic model of rigid link robot manipulators

The Euler–Lagrange equations are a tool from analytical mechanics that can be used to derive the equations of motion for a mechanical system. In this approach the joint $q(t)$ are considered as generalized coordinates. The kinetic energy of a robot manipulator with n degrees of freedom can be calculated as:

$$\Gamma(\mathbf{q},\dot{\mathbf{q}}) = \frac{1}{2}\dot{\mathbf{q}}(t)^T \mathbf{D}(\mathbf{q})\dot{\mathbf{q}}(t),$$

where $D(q)$ is the inertia matrix. Let $U(q): \Re^n \to \Re$ be a continuously differentiable function, called the potential energy. The Lagrangian function is defined (Spong M. W. & Vidyasagar M. 1989) by:

$$L(\mathbf{q},\dot{\mathbf{q}}) = \Gamma(\mathbf{q},\dot{q}) - U(\mathbf{q}).$$

The dynamics of the manipulator are described by Lagrange's equations:

$$\frac{d}{dt}\frac{\partial L(\mathbf{q},\dot{\mathbf{q}})}{\partial \dot{\mathbf{q}}_k} - \frac{\partial L(\mathbf{q},\dot{\mathbf{q}})}{\partial \mathbf{q}_k} = \mathbf{u}_k, \quad k=1,\ldots,n,$$

reference signal in a $r_i^{th}$ order Taylor series, $r_i$ is the relative degree of the $i^{th}$ element of the output. Then, the continuous minimization of the predictive tracking errors where $\mathbf{u}_1, \mathbf{u}_2, \ldots \mathbf{u}_n$ represent generalized input torques. Inserting the kinetic energy and the potential energy for the Lagrange $L(\mathbf{q},\dot{\mathbf{q}})$ above leads to the matrix description:

$$\mathbf{D}(\mathbf{q})\ddot{\mathbf{q}} + \mathbf{C}(\mathbf{q},\dot{\mathbf{q}})\dot{\mathbf{q}} + \mathbf{G}(\mathbf{q}) + \mathbf{f}_r = \mathbf{u}_r,$$

where $q(t) \in \Re^n$ is the vector of the generalized coordinates representing the angular joint positions and controlled with the driving torques $\mathbf{u}_r \in \Re^n$, $D(q) \in \Re^{n \times n}$, $D(q)=D(q)^T>0$, is the link inertia matrix, $\mathbf{C}(\mathbf{q},\dot{\mathbf{q}})\dot{\mathbf{q}} \in \Re^n$ is the vector of the coriolis and centripetal torques, $G(q) \in \Re^n$ is the vector of gravitational torques and $f_r$ represents friction torques acting on the joints. This is described in [9], when only the mechanical parts of actuators dynamics are included. The dynamic model of a rigid robot manipulator becomes:

$$\mathbf{M}(\mathbf{q})\ddot{\mathbf{q}} + \mathbf{C}(\mathbf{q},\dot{\mathbf{q}})\dot{\mathbf{q}} + \mathbf{G}(\mathbf{q}) + \mathbf{f} = \mathbf{u}, \quad (1)$$

with:

$$\mathbf{u} = N\mathbf{u}_m, \qquad \mathbf{M}(\mathbf{q}) = \mathbf{D}(\mathbf{q}) + N^2 \mathbf{J}_m = \mathbf{D}(\mathbf{q}) + \mathbf{J}$$

and $\mathbf{f} = \mathbf{f}_r + N\mathbf{f}_m$,

where:

$N$ is the diagonal matrix of the gear ratios.

$u_m$ is the vector of torque supplied by the actuators.

$f_m$ is the vector friction torque acting on the motors.

$J_m$ is the diagonal matrix containing the effective motors' inertia.

It is assumed that the position $q(t)$ is available for measurement.

***Control Objective:*** The desired reference trajectory for the control object to follow is assumed to be available as bounded functions of time in terms of generalized position $q_{ref}(t)$. That is, there exist three positive constants $r_i$, $i=0,1,2$ such that the following inequalities hold:

$$\|\mathbf{q}_{ref}(t)\| \leq r_0, \ \|\dot{\mathbf{q}}_{ref}(t)\| \leq r_1 \text{ and } \|\ddot{\mathbf{q}}_{ref}(t)\| \leq r_2 \quad (2)$$

***State space representation:*** The dynamic equation of n link robot manipulator (1) can be written in the state space representation as:

$$\begin{cases} \dot{\mathbf{x}}_1(t) = \mathbf{x}_2, \\ \dot{\mathbf{x}}_2(t) = \mathbf{f}(\mathbf{x}_1,\mathbf{x}_2) + \mathbf{P}(\mathbf{x}_1)\mathbf{u}(t), \quad (3) \\ \mathbf{y}(t) = \mathbf{x}_1, \end{cases}$$

where $\mathbf{x}(t) = [\mathbf{x}_1 \ \mathbf{x}_2]^T = [\mathbf{q} \ \dot{\mathbf{q}}]^T \in \Re^{2n}$ is the state vector. $u(t) \in \Re^n$ represents the control torque vector and $y(t) \in \Re^n$ is the output vector (position angular). $\mathbf{f}(\mathbf{x}_1, \mathbf{x}_2) = -\mathbf{M}(\mathbf{q})^{-1}(\mathbf{C}(\mathbf{q},\dot{\mathbf{q}})\dot{\mathbf{q}} + \mathbf{G}(\mathbf{q})) \in \Re^n$ and $\mathbf{P}(\mathbf{x}_1) = \mathbf{M}(\mathbf{x}_1)^{-1} \in \Re^{n\times n}$.

**Properties** (Spong M. W. & Vidyasagar M. 1989):
- $P_1$. The matrix $\mathbf{M}(x_1)$ is symmetric definite positive, then there exist two positive constants: $\underline{M}$ and $\overline{M}$ such that: $\underline{M} \le \|\mathbf{M}(x_1)\| \le \overline{M}$.
- $P_2$. $\exists \mu > 0$ such that $\|\mathbf{C}(\mathbf{q},\mathbf{x})\| \le \mu \|\mathbf{x}\|$, $\forall \mathbf{x} \in \Re^n$.
- $P_3$ The vector function $\mathbf{f}(\mathbf{x}_1,\mathbf{x}_2)$ is Lipschitz with respect to $\mathbf{x}_2$. Thus there exists $\kappa > 0$ such that:

$$\|\mathbf{f}(\mathbf{x}_1,\mathbf{x}_2) - \mathbf{f}(\mathbf{x}_1,\dot{\mathbf{q}}_{ref})\| \le \kappa \|\mathbf{x}_2 - \dot{\mathbf{q}}_{ref}\| = \kappa \|\mathbf{e}_2\|,$$
$\forall (\mathbf{x}_1, \mathbf{x}_2) \in \Re^n \times \Re^n$.

## 3. Receding-horizon control law

In the receding-horizon control strategy, the following control problem is solved at each t>0 and $x(t)$:

$$\underset{u(t)}{Min}\, J(\mathbf{x}(t), t, \mathbf{u}(t)) = \underset{u(t)}{Min}\, \frac{1}{2} \int_t^{t+h} [\mathbf{x}(\tau)^T \mathbf{Q}\, \mathbf{x}(\tau) + \mathbf{u}(\tau)^T \mathbf{R}\, \mathbf{u}(\tau)] d\tau \quad (4)$$

subject to the equation (3) and $\mathbf{x}(t+h) = 0$ for some h>0, where $\mathbf{Q}$ is positive definite and $\mathbf{R}$ positive semi-definite. Denote the optimal control to the above problem by $\mathbf{u}^*(\tau)$, $\tau \in [t, t+h]$. The currently applied control is $\mathbf{u}(t)$ set equal to $\mathbf{u}^*(t)$. This process is repeated for every next $t$ for stabilization of the system at the origin. However, to solve a nonlinear dynamic optimization problem with equality constraints is highly computationally intensive, and in many cases it is impossible to be performed within a reasonable time limit. Furthermore, the global optimization solution cannot be guaranteed in each optimization procedure since, in general, it is a non-convex, constrained nonlinear optimization problem.

In order to find the current control that improves tracking error along a fixed interval, the output tracking error $\mathbf{e}(\tau) = \mathbf{q}(\tau) - \mathbf{q}_{ref}(\tau)$ is used instead of the state vector $\mathbf{x}(\tau)$ in the above receding control problem:

$$J(\mathbf{e},\mathbf{u},t) = \frac{1}{2}\int_t^{t+T}(\mathbf{e}^T(\tau)\mathbf{Q}\mathbf{e}(\tau) + \mathbf{u}(\tau)^T\mathbf{R}\mathbf{u}(\tau))d\tau \quad (5)$$

where $\mathbf{Q} \in \Re^{n \times n}$ is positive definite, $\mathbf{R} \in \Re^{n \times n}$ positive semi-definite, T is the predicted tracking horizon.

To avoid the computational burden, we shall approximate the above receding-horizon control problem by Simpson's rule (Atkinson K. E. 1978):

$$J = \frac{1}{2}\int_t^{t+T} L(\tau)\, d\tau$$
$$= \frac{T}{6}\left[L(t) + 4L\left(t + \frac{T}{2}\right) + L(t+T)\right]$$
$$= \frac{h}{3}[L(t) + 4L(t+h) + L(t+2h)]$$

with T=2h is the prediction horizon and

$$L(\tau) = \mathbf{e}^T(\tau)\mathbf{Q}\,\mathbf{e}(\tau) + \mathbf{u}^T(\tau)\mathbf{R}\,\mathbf{u}(\tau).$$

A simple and effective way of predicting the cost function $L(.)$ is to expand the predicted tracking error in a first order Taylor series, in the following way:
$\mathbf{q}(t+h) = \mathbf{q}(t) + h\,\dot{\mathbf{q}}(t)$ and the reference trajectory is predicted as follow: $\mathbf{q}_{ref}(t+h) = \mathbf{q}_{ref}(t) + h\,\dot{\mathbf{q}}_{ref}$.

The predicted tracking error is then given by:

$$\mathbf{e}(t+h) = \mathbf{e}(t) + h\,\dot{\mathbf{e}}(t).$$

Predict $e$(t+2h) by another first-order Taylor series expansion at $e$(t+h) to have:

$$\begin{aligned}\mathbf{e}(t+2h) &= \mathbf{q}(t+2h) - \mathbf{q}_{ref}(t+2h) \\ &= \mathbf{e}(t) + 2h\,\dot{\mathbf{e}}(t) + h^2(\mathbf{f} - \ddot{\mathbf{q}}_{ref}) + h^2\mathbf{P}\,\mathbf{u}(t)\end{aligned} \quad (6)$$

where $\mathbf{q}(t+2h) = \mathbf{q}(t) + 2h\,\dot{\mathbf{q}}(t) + h^2\mathbf{f} + h^2\mathbf{P}\,\mathbf{u}(t)$ and
$\mathbf{q}_{ref}(t+2h) = \mathbf{q}_{ref}(t) + 2h\,\dot{\mathbf{q}}_{ref}(t) + h^2\,\ddot{\mathbf{q}}_{ref}(t)$.

Thus, the performance index (5) can be approximated as:

$$\begin{aligned}J = \frac{h}{3}\Big[&\mathbf{e}^T(t)\mathbf{Q}\,\mathbf{e}(t) + \mathbf{u}^T(t)\mathbf{R}\,\mathbf{u}(t) \\ &+ 4\mathbf{e}^T(t+h)\mathbf{Q}\,\mathbf{e}(t+h) + 4\mathbf{u}^T(t+h)\mathbf{R}\,\mathbf{u}(t+h) \\ &+ \mathbf{e}^T(t+2h)\mathbf{Q}\,\mathbf{e}(t+2h) + \mathbf{u}^T(t+2h)\mathbf{R}\,\mathbf{u}(t+2h)\Big]\end{aligned}$$
$$(7)$$

We can rewrite the performance index (7) in the conventional quadratic form by using the predicted tracking error given above, as:

$$\bar{J} = \frac{3}{2h}J = \frac{1}{2}\mathbf{U}^T\mathbf{\theta}(x)\,\mathbf{U} + \mathbf{G}^T(x)\,\mathbf{U} + m(\mathbf{e},\dot{\mathbf{e}})$$

where $\mathbf{\theta}(x) = \begin{vmatrix} \mathbf{R} + h^4\mathbf{P}^T(\mathbf{x}_1)\,\mathbf{Q}\,\mathbf{P}(\mathbf{x}_1) & 0 & 0 \\ 0 & 4\mathbf{R} & 0 \\ 0 & 0 & \mathbf{R} \end{vmatrix}$ is positive definite matrix, $m(\mathbf{e},\dot{\mathbf{e}})$: terms that are independent of **U**(t) where **U**(t)$^T$= [ **u**(t)$^T$ **u**(t+h)$^T$ **u**(t+2h)$^T$],

$$\mathbf{G}^T(\mathbf{x}) = \left| h^2 (\mathbf{e} + 2h\dot{\mathbf{e}} + h^2 (\mathbf{f} - \ddot{\mathbf{q}}_{ref})^T \mathbf{Q} \mathbf{P}(\mathbf{x}) \quad 0 \quad 0 \right|.$$

The receding-horizon control problem that minimizes the cost function $\bar{J}$ is : $\mathbf{U}(t) = -\boldsymbol{\theta}(\mathbf{x})^{-1} \mathbf{G}(\mathbf{x})$. The applied control signal to nonlinear system at time t is given by:

$$\mathbf{u}(t) = -h^2 \mathbf{M}(\mathbf{x}_1) \left( h^4 \mathbf{Q} + \mathbf{M}(\mathbf{x}_1) \mathbf{R} \mathbf{M}(\mathbf{x}_1) \right)^{-1} \times \\ \mathbf{Q} \left( \mathbf{e} + 2h \dot{\mathbf{e}} + h^2 \left( \mathbf{f}(\mathbf{x}_1, \mathbf{x}_2) - \ddot{\mathbf{q}}_{ref} \right) \right) \quad (9)$$

Note that with **R**=0, the above nonlinear predictive control law leads to the well known computed torque controller.

**4- Stability analysis and robustness issues**

In this section, we will investigate the stability and the robustness of the closed loop system with respect to model uncertainties.

*1- Stability Analysis*
Let $\mathbf{Q} = q\mathbf{I}_n$ and $\mathbf{R} = r\mathbf{I}_n$ (we give the same penalty to all joints), the tracking error of the nonlinear system (3) closed by the nonlinear feedback (9) is given by:

$$\begin{cases} \dot{\mathbf{e}}_1 = \mathbf{e}_2 \\ \dot{\mathbf{e}}_2 = -q h^2 \bar{\mathbf{P}}^{-1} \mathbf{e}_1 - 2q h^3 \bar{\mathbf{P}}^{-1} \mathbf{e}_2 + r \bar{\mathbf{P}}^{-1} \mathbf{M}(\mathbf{x}_1)^2 \times \\ \quad \left( \mathbf{f}(\mathbf{x}_1, \mathbf{x}_2) - \ddot{\mathbf{q}}_{ref} \right) \end{cases}$$

(10)

Where $\mathbf{e}_1 = \mathbf{q}(t) - \mathbf{q}_{ref}(t)$; $\mathbf{e}_2 = \dot{\mathbf{q}}(t) - \dot{\mathbf{q}}_{ref}(t)$ and $\bar{\mathbf{P}} = qh^4 \mathbf{I}_n + r \mathbf{M}(\mathbf{x}_1)^2$.

This equation can be written in compact form as :

$$\dot{\mathbf{e}} = \mathbf{A}(h, \mathbf{x}_1) \mathbf{e} + \mathbf{B} \boldsymbol{\eta} \quad (11)$$

Where $\mathbf{A}(h, \mathbf{x}_1) = \left| \begin{matrix} 0 & \mathbf{I}_n \\ -qh^2 \bar{\mathbf{P}}^{-1} & -2qh^3 \bar{\mathbf{P}}^{-1} \end{matrix} \right|$; $\mathbf{B} = \left| \begin{matrix} 0 \\ \mathbf{I}_n \end{matrix} \right|$;

$\boldsymbol{\eta} = r \bar{\mathbf{P}}^{-1} \mathbf{M}(\mathbf{x}_1)^2 \left( \mathbf{f}(\mathbf{x}_1, \mathbf{x}_2) - \ddot{\mathbf{q}}_{ref} \right).$

**Lemma-1**: The matrix **A**(h,**x**$_1$) is Hurwitz.

**Proof**: Both matrix $\bar{\mathbf{P}}$ and its inverse are symmetric positives definite. Let $\bar{\mathbf{x}} \in \mathfrak{R}^n$ and $\bar{\lambda} \in \mathfrak{R}$ are the eigenvector and the correspondent eigenvalue of the inverse of the above matrix. Thus, we have the equality:

$$\mathbf{A}(h, \mathbf{x}_1) \left| \begin{matrix} \bar{\mathbf{x}} \\ \lambda \bar{\mathbf{x}} \end{matrix} \right| = \left| \begin{matrix} \lambda \bar{\mathbf{x}} \\ -qh^2 \bar{\lambda} \bar{\mathbf{x}} - 2qh^3 \lambda \bar{\lambda} \bar{\mathbf{x}} \end{matrix} \right| = \left| \begin{matrix} \lambda \bar{\mathbf{x}} \\ \lambda^2 \bar{\mathbf{x}} \end{matrix} \right| = \lambda \left| \begin{matrix} \bar{\mathbf{x}} \\ \lambda \bar{\mathbf{x}} \end{matrix} \right| \text{N}$$

ote that $\lambda$ is the solution of equation:

$$\lambda^2 + qh^2 \bar{\lambda} + 2qh^3 \bar{\lambda} \lambda = 0 \quad (12)$$

Therefore, $\lambda$ is the eigenvalue of the matrix **A**(h, **x**$_1$) and $\left| \begin{matrix} \bar{\mathbf{x}} \\ \lambda \bar{\mathbf{x}} \end{matrix} \right|$ the correspondent eigenvector. Set $\lambda_1$ and $\lambda_2$ the solution of the equation (12), we have the relations:

$$\lambda_1 + \lambda_2 = -2qh^3 \bar{\lambda}$$
$$\lambda_1 \lambda_2 = qh^2 \bar{\lambda}$$

Since the eigenvalue $\bar{\lambda}$ is positive, then $\lambda_1$ and $\lambda_2$ have a negative real part (end of the proof).

Since the matrix **A**(h,**x**$_1$) is a Hurwitz matrix, then for any symmetric positive definite matrix **Q$_A$**(h, **x**$_1$), there exists a symmetric positive definite matrix **P$_A$**(h, **x**$_1$) solution of the lyapunov equation:

$$\dot{\mathbf{P}}_\mathbf{A}(h, \mathbf{x}_1) + \mathbf{A}(h, \mathbf{x}_1)^T \mathbf{P}_\mathbf{A}(h, \mathbf{x}_1) + \mathbf{P}_\mathbf{A}(h, \mathbf{x}_1) \mathbf{A}(h, \mathbf{x}_1) \\ = -\mathbf{Q}_\mathbf{A}(h, \mathbf{x}_1)$$

(13)

From the property P$_3$, the function $\mathbf{f}(\mathbf{x}_1, \mathbf{x}_2)$ is Lipschitz with regards to **x**$_2$, we can always find a bounded continuous function $\sigma(\mathbf{e}_2, t)$ and positive scalar $\mu$ satisfying the inequality:

$$\left\| \mathbf{f}(\mathbf{x}_1, \mathbf{x}_2) - \ddot{\mathbf{q}}_{ref} \right\| \leq \sigma(\mathbf{e}_2, t) \leq \mu \|\mathbf{e}\|. \quad (14)$$

Now, we can state the following theorem.

**Theorem 1**: The equilibrium point of the nonlinear system (3) in closed loop with the feedback control (9) is asymptotically stable if the following inequality hold:

$$\mu < \frac{\lambda_{min}(\mathbf{Q}_\mathbf{A}) \lambda_{max}(\bar{\mathbf{P}})}{2 r \bar{M}^2 \lambda_{max}(\mathbf{P}_\mathbf{A})}.$$

Moreover, if r = 0, then the origin is asymptotically stable equilibrium point.

**Proof:**
Let $V(\mathbf{e}) = \mathbf{e}^T \mathbf{P}_A \mathbf{e}$ be a Lyapunov function candidate, the time derivative of this function along the trajectories (11) is:

$$\dot{V} = \mathbf{e}^T \mathbf{A}(h, \mathbf{x}_1)^T \mathbf{P}_A \mathbf{e} + \mathbf{e}^T \mathbf{P}_A \mathbf{A}(h, \mathbf{x}_1) \mathbf{e} + \mathbf{e}^T \dot{\mathbf{P}}_A \mathbf{e} \\ + \boldsymbol{\eta}^T \mathbf{B}^T \mathbf{P}_A \mathbf{e} + \mathbf{e}^T \mathbf{P}_A \mathbf{B} \boldsymbol{\eta}$$

By using the equality (13), we obtain:

$$\dot{V} = -\mathbf{e}^T \mathbf{Q}_A \mathbf{e} + \mathbf{\eta}^T \mathbf{B}^T \mathbf{P}_A \mathbf{e} + \mathbf{e}^T \mathbf{P}_A \mathbf{B} \mathbf{\eta}$$

or $\quad \dot{V} \leq -\lambda_{min}(\mathbf{Q}_A) \|\mathbf{e}\|^2 + 2 \|\mathbf{\eta}\| \|\mathbf{B}\| \|\mathbf{P}_A\| \|\mathbf{e}\|$

From the inequality (14), we can write:

$$\dot{V} \leq -\lambda_{min}(\mathbf{Q}_A) \|\mathbf{e}\|^2 + 2\mu r \overline{M}^2 \lambda_{max}(\overline{\mathbf{P}}^{-1}) \lambda_{max}(\mathbf{P}_A) \|\mathbf{e}\|^2.$$

Thus, $\dot{V} \leq -\kappa \|\mathbf{e}\|^2$ is negative definite if $\kappa > 0$, where $\kappa = \lambda_{min}(\mathbf{Q}_A) - 2\mu \overline{M}^2 r \dfrac{\lambda_{max}(\mathbf{P}_A)}{\lambda_{max}(\overline{\mathbf{P}})}$. This ensures the asymptotic stability of the equilibrium point.

Note that a short steady state error will be observed in tracking position error when $r \neq 0$. However, if r =0 the time derivative of the Lyapunov function becomes: $\dot{V} = -\lambda_{min}(\mathbf{Q}_A) \|\mathbf{e}\|^2$ which is negative definite. Thus, we can conclude that the origin becomes the equilibrium point of the system (12) and is asymptotically stable, i.e:

$$\underset{t\to\infty}{Lim}\,\mathbf{e}(t) = \underset{t\to\infty}{Lim}\,\left|\mathbf{q}-\mathbf{q}_{ref}\ \ \dot{\mathbf{q}}-\dot{\mathbf{q}}_{ref}\right|^T = \left|0\ \ 0\right|^T.$$

## 2- Robustness

In order to incorporate modeling uncertainties into the model of the rigid robot (1), the matrices $\mathbf{M}(\mathbf{q})$, $\mathbf{C}(\mathbf{q},\dot{\mathbf{q}})$ and the vector $\mathbf{G}(\mathbf{q})$ are split up into a nominal part (indicated by the subscript zero) and an uncertain part as:

$$(\mathbf{M}_0(\mathbf{q}) + \Delta\mathbf{M})\ddot{\mathbf{q}} + (\mathbf{C}_0(\mathbf{q},\dot{\mathbf{q}}) + \Delta\mathbf{C})\dot{\mathbf{q}} + \mathbf{G}_0(\mathbf{q}) + \Delta\mathbf{G} + \mathbf{f}$$
$$= \mathbf{u}(t) \quad (15)$$

With $\mathbf{M}(\mathbf{q}) = \mathbf{M}_0(\mathbf{q}) + \Delta\mathbf{M}$; $\mathbf{C}(\mathbf{q},\dot{\mathbf{q}}) = \mathbf{C}_0(\mathbf{q},\dot{\mathbf{q}}) + \Delta\mathbf{C}$ and $\mathbf{G}(\mathbf{q}) = \mathbf{G}_0(\mathbf{q}) + \Delta\mathbf{G}$.

The friction torque $f$ is included in the uncertain part given the difficulty to model it correctly. Obviously, only the nominal part of the model can be used by the nonlinear predictive control, given by:

$$\mathbf{u}(t) = -\frac{1}{h^2}\mathbf{M}_0(q)(\mathbf{e} + 2h\dot{\mathbf{e}}) \\ + (\mathbf{C}_0(\mathbf{q},\dot{\mathbf{q}})\dot{\mathbf{q}} + \mathbf{G}_0(\mathbf{q})) + \mathbf{M}_0(\mathbf{q})\ddot{\mathbf{q}}_{ref} \quad (16)$$

Where $\mathbf{R}$ is setting to zero. With the nonlinear control law (16), the closed loop system is:

$$\ddot{\mathbf{e}} + \frac{2}{h}\mathbf{M}(\mathbf{q})^{-1}\mathbf{M}_0(\mathbf{q})\dot{\mathbf{e}} + \frac{1}{h^2}\mathbf{M}(\mathbf{q})^{-1}\mathbf{M}_0(\mathbf{q})\mathbf{e}$$
$$= -\mathbf{M}^{-1}(\mathbf{q})(\Delta\mathbf{M}\ddot{\mathbf{q}}_{ref} + \Delta\mathbf{C}\dot{\mathbf{q}} + \Delta\mathbf{G} + \mathbf{f}) = \mathbf{\upsilon}(\mathbf{q},\dot{\mathbf{q}},\ddot{\mathbf{q}}_{ref},t)$$
(17)

To estimate the worst case bound of the function $\upsilon$, we make the following assumptions for all $\mathbf{q} \in \Re^n$:

$\underline{M}_0 \leq \|\mathbf{M}_0(\mathbf{q})\| \leq \overline{M}_0$;  $\quad \|\Delta\mathbf{M}\| \leq \overline{m} \leq \lambda_{\min}(M(\mathbf{q}))$;

$\|\Delta\mathbf{C}\| \leq \overline{c}$;  $\quad \|\Delta\mathbf{G}\| \leq \overline{g}$;  $\quad \|f\| \leq \overline{f}$;

Given these assumptions with the inequalities (2), we can find a bounded continuous vector function $\rho(\mathbf{e},\dot{\mathbf{e}},t)$ satisfying the inequality (Spong M. W. & Vidyasagar M.;1989):

$$\|\upsilon\| < \rho(\mathbf{e},\dot{\mathbf{e}},t) \leq \gamma \|\mathbf{e}\| \quad \text{for all } \mathbf{q} \in \Re^n,$$

where $\gamma$ is a positive scalar and $\mathbf{e} = |\mathbf{q}\ \ \dot{\mathbf{q}}|^T = |\mathbf{e}_1\ \ \mathbf{e}_2|^T$.

In the state space representation, the system (17) can be transformed to:

$$\dot{\mathbf{e}} = \overline{\mathbf{B}}(h,\mathbf{x}_1)\mathbf{e} + \mathbf{B}\upsilon \quad (18)$$

where $\overline{\mathbf{B}}(h,\mathbf{x}_1) = \begin{vmatrix} 0 & \mathbf{I}_n \\ -\dfrac{\mathbf{b}(\mathbf{x}_1)}{h^2} & -\dfrac{2\mathbf{b}(\mathbf{x}_1)}{h} \end{vmatrix}$.

Since both $\mathbf{M}^{-1}(\mathbf{x}_1)$ and $\mathbf{M}_0(\mathbf{x}_1)$ are symmetric definite positives, the matrix $\mathbf{b}(\mathbf{x}_1) = \mathbf{M}^{-1}(\mathbf{x}_1)\mathbf{M}_0(\mathbf{x}_1)$ has all its eigenvalues reels and positives (Samson C.;1983). Thus, from the Lemma-1, we can conclude that the matrix $\overline{\mathbf{B}}(h,\mathbf{x}_1)$ is Hurwitz, then for any symmetric positive definite matrix $\mathbf{Q}_B(h,\mathbf{x}_1)$ there exists a positive definite matrix $P_B(h,x_1)$ solution of the given Lyapunov equation:

$$\dot{\mathbf{P}}_B(h,\mathbf{x}_1) + \overline{\mathbf{B}}(h,\mathbf{x}_1)^T \mathbf{P}_B(h,\mathbf{x}_1) + \mathbf{P}_B(h,\mathbf{x}_1)\overline{\mathbf{B}}(h,\mathbf{x}_1) \\ = -\mathbf{Q}_B(h,\mathbf{x}_1)$$

**Theorem 2**: Suppose that the inequality holds $\gamma < \dfrac{\lambda_{min}(\mathbf{Q}_B)}{2\lambda_{max}(\mathbf{P}_B)}$, then the equilibrium point of the nonlinear system with uncertainties (15) closed by the optimal control (16) is asymptotically stable.

**Proof:**
Let $V = \mathbf{e}^T \mathbf{P}_B \mathbf{e}$ a Lyapunov function candidate. The differentiation of $V$ along the trajectories (18) leads to:

$$\dot{V} = -\mathbf{e}^T \mathbf{Q}_B \mathbf{e} + \upsilon^T \mathbf{B}^T \mathbf{P}_B \mathbf{e} + \mathbf{e}^T \mathbf{P}_B \mathbf{B}\upsilon.$$
$$\dot{V} \leq -\lambda_{min}(\mathbf{Q}_B)\|\mathbf{e}\|^2 + 2\gamma\lambda_{max}(\mathbf{P}_B)\|\mathbf{e}\|^2$$
$$\leq -(\lambda_{min}(\mathbf{Q}_B) - 2\gamma\lambda_{max}(\mathbf{P}_B))\|\mathbf{e}\|^2.$$

Which is definite negative if $\gamma < \dfrac{\lambda_{min}(\mathbf{Q}_B)}{2\lambda_{max}(\mathbf{P}_B)}$.

Therefore, By LaSalle's invariance theorem, the solution $\mathbf{e}(t)$ of (18) tends to the invariance set :

$$S = \{e \,/\, \mathbf{e}_1 = h^2 \mathbf{b}(\bar{\mathbf{x}}_1)^{-1} \mathbf{v}(\bar{\mathbf{x}}, \ddot{\mathbf{q}}ref), \mathbf{e}_2 = 0\}.$$

We conclude that bounded uncertainties will introduce a steady state error on tracking position angular. Where $\bar{\mathbf{x}}$ is the equilibrium point of the system (3) in closed loop with the control law (16).

### 3- Integral action

It is known in the literature that the integral action increases the robustness of the closed loop system against the low frequency disturbances as long as the closed loop system is stable. In this part, we shall incorporate an integral action in the loop to eliminate the steady state error and enhance the robustness of the proposed control scheme with respect to model uncertainties and disturbances.

Thus, the cost function to minimize becomes:

$$J(\mathbf{e}_0,\mathbf{u},t) = \frac{1}{2} \int_t^{t+T} \left( \mathbf{e}_0^T(\tau) \mathbf{Q} \mathbf{e}_0(\tau) + \mathbf{u}(\tau)^T \mathbf{R} \mathbf{u}(\tau) \right) d\tau$$
$$= \frac{1}{2} \int_t^{t+h} L(\tau) d\tau \qquad (19)$$

Where $\dot{\mathbf{e}}_0 = \mathbf{e}_1 = \mathbf{x}_1 - \mathbf{q}_{ref} = \mathbf{q} - \mathbf{q}_{ref}$

and $L(\tau) = \mathbf{e}_0^T(\tau) \mathbf{Q} \mathbf{e}_0(\tau) + \mathbf{u}^T(\tau) \mathbf{R} \mathbf{u}(\tau)$.

Also in this case, we use the Simpson's rule to approximate the integral in the cost function (19) by:

$$J(\mathbf{e}_0, \mathbf{u}, t) = \frac{h}{3}(L(t) + 4L(t+h) + L(t+2h))$$

With: $L(t) = \mathbf{e}_0^T(t) \mathbf{Q} \mathbf{e}_0(t) + \mathbf{u}^T(t) \mathbf{R} \mathbf{u}(t)$;

$L(t+h) = \mathbf{e}_0^T(t+h) \mathbf{Q} \mathbf{e}_0(t+h) + \mathbf{u}^T(t+h) \mathbf{R} \mathbf{u}(t+h)$

$L(t+2h) = \mathbf{e}_0^T(t+2h) \mathbf{Q} \mathbf{e}_0(t+2h)$
$\qquad + \mathbf{u}(t+2h) \mathbf{R} \mathbf{u}(t+2h)$.

Note that in this case, the Taylor approximation of the predicted vector $\mathbf{e}_0(t+h)$ is given by:

$$\mathbf{e}_0(t+h) = \mathbf{e}_0(t) + h\mathbf{e}_1 + \frac{h^2}{2}\mathbf{e}_2 + \frac{h^3}{6}(\mathbf{f}(x) - \ddot{\mathbf{q}}_{ref})$$
$$+ \frac{h^3}{6} \mathbf{P}(\mathbf{x}_1) \mathbf{u}(t)$$

Following same steps in paragraph 2, the optimal control vector $\mathbf{U}(t)$ that minimizes the new cost function is:

$$\mathbf{U}(t) = -\overline{\mathbf{H}}^{-1} \overline{\mathbf{G}}(\mathbf{x}) \qquad (20)$$

Where $\mathbf{U}(t) = |\mathbf{u}(t) \quad \mathbf{u}(t+h) \quad \mathbf{u}(t+2h)|^T$;

$\overline{\mathbf{H}} = diag(\mathbf{R} + \frac{5}{9} h^6 \mathbf{P}^T(\mathbf{x}_1) \mathbf{Q} \mathbf{P}(\mathbf{x}_1), 4\mathbf{R}, \mathbf{R})$ and

$$\overline{\mathbf{G}}(\mathbf{x}) = \left| \begin{array}{c} \frac{2}{3} h^3 \mathbf{P}(\mathbf{x}_1) \mathbf{Q} \left( 2\mathbf{e}_0 + 3h\mathbf{e}_1 + 2h^2\mathbf{e}_2 + \frac{5}{6} h^3 (\mathbf{f}(\mathbf{x}) - \ddot{\mathbf{q}}_{ref}) \right) \\ 0 \\ 0 \end{array} \right|$$

The control signal to be applied to the nonlinear system at time $t$ is:

$$\mathbf{u}(t) = -\frac{2}{3} h^3 \mathbf{M}_0(\mathbf{x}_1) \overline{\overline{\mathbf{P}}}^{-1}(h, \mathbf{x}_1) \mathbf{Q} \times$$
$$\left( 2\mathbf{e}_0 + 3h\mathbf{e}_1 + 2h^2\mathbf{e}_2 + \frac{5}{6} h^3 (\mathbf{f}(\mathbf{x}) - \ddot{\mathbf{q}}_{ref}) \right) \qquad (21)$$

Where $\overline{\overline{\mathbf{P}}}(h, \mathbf{x}_1) = \frac{5}{9} h^6 \mathbf{Q} + \mathbf{M}_0(\mathbf{x}_1) \mathbf{R} \mathbf{M}_0(\mathbf{x}_1)$.

Let $\mathbf{R} = 0$ in equation (21), the control signal becomes:

$$\mathbf{u}(t) = -\frac{9}{5} \mathbf{M}_0(\mathbf{x}_1) \times$$
$$\left( \frac{4}{3h^3} \mathbf{e}_0 + \frac{2}{h^2} \mathbf{e}_1 + \frac{4}{3h} \mathbf{e}_2 + \frac{5}{9} (\mathbf{f}(\mathbf{x}) - \ddot{\mathbf{q}}_{ref}) \right)$$

The dynamic of the tracking error is given by the equations:

$$\begin{cases} \dot{\mathbf{e}}_0 = \mathbf{e}_1 \\ \dot{\mathbf{e}}_1 = \mathbf{e}_2 \\ \dot{\mathbf{e}}_2 = -\frac{12}{5h^3} \mathbf{b}(\mathbf{x}_1) \mathbf{e}_0 - \frac{18}{5h^2} \mathbf{b}(\mathbf{x}_1) \mathbf{e}_1 - \frac{12}{5h} \mathbf{b}(\mathbf{x}_1) \mathbf{e}_2 \\ \qquad + \mathbf{v}(\mathbf{x}, \ddot{\mathbf{q}}_{ref}) \end{cases}$$
(22)

Or in compact form:

$$\dot{\mathbf{e}} = \widetilde{\mathbf{B}}(h, \mathbf{x}_1) \mathbf{e} + \mathbf{B} \, \mathbf{v}(\mathbf{x}, \ddot{\mathbf{q}}_{ref}) \qquad (23)$$

Where:

$$\widetilde{\mathbf{B}}(h, x_1) = \left| \begin{array}{ccc} 0 & I_n & 0 \\ 0 & 0 & I_n \\ -\frac{12}{5h^3} \mathbf{b}(\mathbf{x}_1) & -\frac{18}{5h^2} \mathbf{b}(\mathbf{x}_1) & -\frac{12}{5h} \mathbf{b}(\mathbf{x}_1) \end{array} \right|$$

**Lemma 2**: Suppose that $\lambda_{max}(\varepsilon(\mathbf{x}_1)) < 2.6$, with $\varepsilon(\mathbf{x}_1) = \mathbf{M}_0^{-1} \Delta \mathbf{M}$, then the matrix $\overline{\overline{\mathbf{B}}}(h, \mathbf{x}_1)$ is Hurwitz.

**Proof:** Let $\bar{\mathbf{x}} \in \mathfrak{R}^n$ and $\bar{\lambda} \in \mathfrak{R}^+$ represent eigenvector and eigenvalue of the matrix $\mathbf{b}(\mathbf{x}_1)$ respectively. Set $\mathbf{v} = \left| \bar{\mathbf{x}} \quad \widetilde{\lambda} \bar{\mathbf{x}} \quad \widetilde{\lambda}^2 \bar{\mathbf{x}} \right|^T \in \mathfrak{R}^{3n}$ be a vector and $\widetilde{\lambda} \in \mathfrak{R}$ a scalar. We have:

$$\widetilde{\mathbf{B}}(h,\mathbf{x}_1)\,\mathbf{v} = \begin{vmatrix} \widetilde{\lambda}\,\overline{\mathbf{x}} \\ \widetilde{\lambda}^2\,\overline{\mathbf{x}} \\ -\dfrac{12}{5h^3}\overline{\lambda}\,\overline{\mathbf{x}} - \dfrac{18}{5h^2}\overline{\lambda}\,\widetilde{\lambda}\,\overline{\mathbf{x}} - \dfrac{12}{5h}\widetilde{\lambda}^2\,\overline{\lambda}\,\overline{\mathbf{x}} \end{vmatrix}$$

$$= \begin{vmatrix} \widetilde{\lambda}\,\overline{\mathbf{x}} \\ \widetilde{\lambda}^2\,\overline{\mathbf{x}} \\ \widetilde{\lambda}^3\,\overline{\mathbf{x}} \end{vmatrix} = \widetilde{\lambda}\,\mathbf{v}$$

Then, $\widetilde{\lambda}$ and $\mathbf{v}$ are the eigenvalue and eigenvector of the matrix $\widetilde{\mathbf{B}}(h,\mathbf{x}_1)$ respectively, where the eigenvalue $\widetilde{\lambda}$ verify the equality:

$$\widetilde{\lambda}^3 + \frac{12}{5h}\overline{\lambda}\,\widetilde{\lambda}^2 + \frac{18}{5h^2}\overline{\lambda}\,\widetilde{\lambda} + \frac{12}{5h^3}\overline{\lambda} = 0.$$

By using the Rooth-Hurwitz criterion, the solutions of the above equation lie in left half plane (stable domain) if: $\overline{\lambda} > \dfrac{5}{18}$.

Since $\mathbf{M}^{-1}\mathbf{M}_0 = (\mathbf{I}_n + \boldsymbol{\varepsilon}(\mathbf{x}_1))^{-1}$, the condition for stability becomes: $\lambda_{max}(\mathbf{I}_n + \boldsymbol{\varepsilon}(\mathbf{x}_1)) < \dfrac{18}{5}$ or $\lambda_{max}(\boldsymbol{\varepsilon}(\mathbf{x}_1)) < 2.6$ which can be easily verified if the uncertainty $\Delta\mathbf{M}$ is small with regards to the nominal value of the matrix $\mathbf{M}_0(\mathbf{x}_1)$.

We conclude that the matrix $\widetilde{\mathbf{B}}(h,\mathbf{x}_1)$ is Hurwitz, then for any symmetric positive definite matrix $\widetilde{\mathbf{Q}}(h,\mathbf{x}_1)$ there exists a positive definite matrix $\widetilde{\mathbf{P}}(h,\mathbf{x}_1)$ solution of the given Lyapunov equation:

$$\dot{\widetilde{\mathbf{P}}}(h,\mathbf{x}_1) + \widetilde{\mathbf{B}}(h,\mathbf{x}_1)^T\,\widetilde{\mathbf{P}}(h,\mathbf{x}_1) + \widetilde{\mathbf{P}}(h,\mathbf{x}_1)\,\widetilde{\mathbf{B}}(h,\mathbf{x}_1)$$
$$= -\widetilde{\mathbf{Q}}(h,\mathbf{x}_1)$$

**Theorem 3**: Suppose the inequality $\gamma < \dfrac{\lambda_{min}(\widetilde{\mathbf{Q}})}{2\,\lambda_{max}(\widetilde{\mathbf{P}})}$ hold, then the equilibrium point of the nonlinear system with uncertainties (23) is asymptotically stable.

The proof may be obtained in the same way as the proof of the theorem 2 and is therefore omitted. Therefore, the equilibrium point of (22) or (23) is asymptotically stable. The tracking error tends towards the set:

$$S = \left\{ \mathbf{e}\,/\,\mathbf{e}_0 = \frac{5h^3}{12}\mathbf{b}(\overline{\mathbf{x}}_1)^{-1}\,\mathbf{v}(\overline{\mathbf{x}},\ddot{q}_{ref}),\,\mathbf{e}_1 = 0,\,\mathbf{e}_2 = 0 \right\} W$$

e conclude that the position and velocity tracking error converge to zero, therefore the integral action eliminates the position steady state error, i.e:

$$\underset{t\to\infty}{Lim}\,\mathbf{e}_1 = 0 \text{ and } \underset{t\to\infty}{Lim}\,\mathbf{e}_2 = 0.$$

The price to be paid by introducing an integral action in the loop is that the control signal will not vanish and this will increase the required energy to maintain the tracking performance as in the matched case.

## 5. Nonlinear observer

A drawback of the previous nonlinear predictive controller is that it requires at least the measurement of velocity on the link side. However, as pointed out in (Nicosia S. & Tomei P. 1990) and (Canudas W. C. *et al* 1992), in the practical robotic systems all the generalized coordinates can be precisely measured by the encoder for each joint, but the velocity measurements obtained through the tachometers are easily perturbed by noises. Therefore, in order to coincide with these physical constraints, a nonlinear observer proposed in (Bornard G. *et al* 1993) is used in this paper.

Define the state vector as:

$$\mathbf{z}(t) = \mathbf{T}\,\mathbf{x}(t) = [......q_i(t)\,\dot{q}_i(t).....] \in \Re^{2n}$$

where $q_i(t)$ and $\dot{q}_i(t)$ are the link position and the velocity of the $i^{th}$ arm respectively. $\mathbf{T} \in \Re^{2n \times 2n}$ is the transformation matrix. The system (3) can be transformed to:

$$\begin{cases} \dot{\mathbf{z}} = \mathbf{A}\,\mathbf{z} + \mathbf{H}\,\mathbf{f}(\mathbf{z}) + \mathbf{H}\,\mathbf{P}(\mathbf{q})\,\mathbf{u}(t) \\ \mathbf{y} = \mathbf{C}\,\mathbf{z} \end{cases} \quad (24)$$

where

$\mathbf{A}=\text{diag}(\mathbf{A}_i)$, $\mathbf{A}_i = \begin{vmatrix} 0 & 1 \\ 0 & 0 \end{vmatrix}$, $\mathbf{C}=\text{diag}(\mathbf{C}_i)$, $\mathbf{C}_i = \begin{vmatrix} 1 & 0 \end{vmatrix}$,

$\mathbf{H}=\text{diag}(\mathbf{H}_i)$, $\mathbf{H}_i = \begin{vmatrix} 0 & 1 \end{vmatrix}^T$ for i=1,......, n.

With the assumption that the control torque $\mathbf{u}(t)$ is uniformly bounded, the high gain observer described in (Bornard G. *et al*, 1993) can be used to estimate angular positions and angular velocities of the n link rigid robot manipulator (24). The dynamic nonlinear observer is given by:

$$\begin{cases} \dot{\hat{\mathbf{z}}} = \mathbf{A}\,\hat{\mathbf{z}} + \mathbf{H}\,\mathbf{f}(\hat{\mathbf{z}}) + \mathbf{H}\,\mathbf{P}(\mathbf{q})\,\mathbf{u}(t) + \mathbf{K}(\mathbf{y}-\hat{\mathbf{y}}), \\ \hat{\mathbf{y}} = \mathbf{C}\,\hat{\mathbf{z}}, \end{cases} \quad (25)$$

where $\mathbf{K} = \boldsymbol{\Gamma}^{-1}(\alpha)\,\mathbf{V}$ is the gain of the observer with $\boldsymbol{\Gamma}(\alpha) = diag(\boldsymbol{\Gamma}_i(\alpha))$, $\boldsymbol{\Gamma}_i(\alpha) = \begin{vmatrix} \alpha & 0 \\ 0 & \alpha^2 \end{vmatrix}$ for any $\alpha > 0$ and due to the observability property of $(\mathbf{A},\mathbf{C})$ the eigenvalues of $(\mathbf{A}-\mathbf{V}\mathbf{C})$ can be assigned by $\mathbf{V}$.

## 6. A Simulation example

To illustrate some of the conclusions of this paper, we have simulated the approximate receding-horizon control scheme on a two-link robot arm used in (Lee K. W. & Khalil H. K. 1997), (Spong M. W. 1992) with mechanic equations.
The arm is shown in Figure 1. The dynamic model is described in equation (1) with the following components:

$$m_{11}(\mathbf{q}) = m_1 l_{c1}^2 + m_2 l_{c2}^2 + m_2 l_1^2$$
$$+ 2 m_2 l_1 l_{c2} \cos(q_2) + I_1 + I_2 \; ;$$
$$m_{12}(\mathbf{q}) = m_{21}(\mathbf{q}) = I_2 + m_2 l_{c2}^2 + m_2 l_1 l_{c2} \cos(q_2) \; ;$$
$$m_{22}(\mathbf{q}) = I_2 + m_2 l_{c2}^2 \; ; \; C_{11}(\mathbf{q},\dot{\mathbf{q}}) = -\dot{q}_2 m_2 l_1 l_{c2} \sin(q_2) \; ;$$
$$C_{12}(\mathbf{q},\dot{\mathbf{q}}) = -(\dot{q}_1 + \dot{q}_2) m_2 l_1 l_{c2} \sin(q_2) \; .$$
$$C_{21}(\mathbf{q},\dot{\mathbf{q}}) = \dot{q}_1 m_2 l_1 l_{c2} \sin(q_2) \; , \; C_{22}(\mathbf{q},\dot{\mathbf{q}}) = 0 \; .$$
$$G_1(\mathbf{q}) = (m_1 l_{c1} + m_2 l_1) g \cos(q_1) + m_2 l_{c2} \; g \cos(q_1 + q_2)$$
$$G_2(\mathbf{q}) = m_2 l_{c2} \; g \cos(q_1 + q_2) \; .$$

where $l_{c1}$ is the mass center of gravity coordinate of the link 1, and $l_{c2}$ is the mass center of gravity coordinate of the link 2. The values of the manipulator parameters are given in Table 1 (Lee K. W. & Khalil H. K. 1997) and (Spong M. W. 1992).

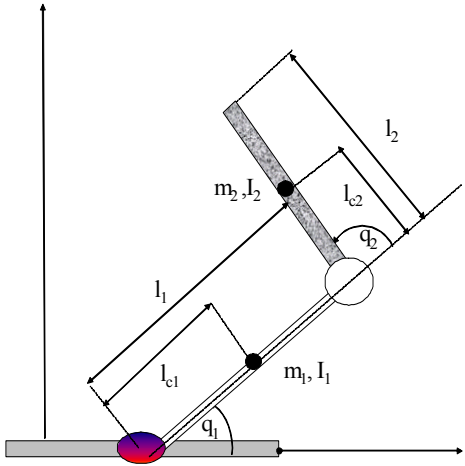

Fig.1 Two-linkage manipulators

| Link 1 | $m_1$=10kg | $l_1$=1m | $l_{c1}$=0.5m |
|---|---|---|---|
| | $I_1 = \dfrac{10}{12}$ kgm$^2$ | | |
| Link 2 | $m_2$=5kg | $l_2$ = 1m | |
| | $l_{c2}$=0.50m | $I_2 = \dfrac{5}{12}$ kgm$^2$ | |

Table 1 Physical parameters of two-link arm.

The reference models chosen in continuous time are:

$$\mathbf{q}_{ref} = \begin{vmatrix} q_{ref1} \\ q_{ref2} \end{vmatrix} \quad \text{with} \quad q_{ref1}(s) = \frac{\omega_1^2}{s^2 + 2\xi\omega_1 s + \omega_1^2} r_1(s)$$

and $q_{ref2}(s) = \dfrac{\omega_2^2}{s^2 + 2\xi\omega_2 s + \omega_2^2} r_2(s)$.

The nonlinear predictive controller is used to track these desired trajectories with inputs (Lee K. W. & Khalil H. K. 1997):

$$r_1(t) = r_2(t) = 1.5 \left(1 - \exp(-5t)(1+5t)\right) rad \; .$$

All simulations are carried with the nonlinear observer (25) with α=0.01 and the assigned eigenvalues $\sigma(\lambda) = \{-0.4, -0.8\}$. The initial displacements and velocities are chosen as:

$q_1(0)=q_2(0)=0°$, $\dot{q}_1(0) = \dot{q}_2(0) = 0$,
$\hat{q}_1(0) = \hat{q}_2(0) = 0.01$ and $\dot{\hat{q}}_1(0) = \dot{\hat{q}}_2(0) = 0$ .

The parameter values of two-reference models are chosen as follows: $\xi=1, \omega_1=\omega_2=10 \; rad/s$.

The nonlinear controller (9), has been tested by simulation and the control parameters: $\mathbf{Q}=10^7 \; \mathbf{I}_n$, $\mathbf{R}=10^{-14} \; \mathbf{I}_n$ and h is set to 0.001. Simulation results are show in Figure 2. This Figure gives the angular position ($q_1(t)$, $q_2(t)$) and the position tracking error. Although a very short steady state error is observed in the position tracking error and this was expected in the analysis part, a good tracking performance is achieved by this controller in matched case. Figure 3 illustrates the induced control torque applied to robot manipulator. Note that the control torque lie inside the saturation limits (Lee K. W. & Khalil H. K. 1997).
In mismatched case, the frictions are added to the joint of robot manipulator model in equation (1) and are modeled: $\mathbf{F}_r = \mathbf{f}_s \; \mathbf{q} + \mathbf{f}_v \; sign(\dot{\mathbf{q}})$, with the values $\mathbf{f}_s = \mathbf{f}_v = diag(5,5)$.

If we regard an unknown load carried by the robot as part of the second link, then the parameters $m_2$, $l_{c2}$ and $I_2$ will change to: $m_2 + \Delta m_2$, $l_{c2} + \Delta l_{c2}$ and $I_2 + \Delta I_2$.

Let $\Delta m_2 = 5Kg$, $\Delta l_{c2} = 0.5m$ and and $\Delta I_2 = \dfrac{1}{6} Kgm^2$ to be the maximum parameters variations of the second link due to unknown load . It is observed from Figure 4 that the output $q_2(t)$ tracks tightly the reference trajectory with a steady state error. This results has been expected in the analysis did in section 4, i.e the uncertainties will introduce a steady state error in tracking error. Furthermore, the induced torque control lie outside the saturation limits. Figure 5 shows the results when the control law (21) is applied to robot, it is seen from this figure the error was eliminated and the torque control signal lie inside the saturation limits. These results prove the robustness of the rigid link manipulator under the approximate receding-horizon controller with integral action to payload uncertainties and viscous friction.

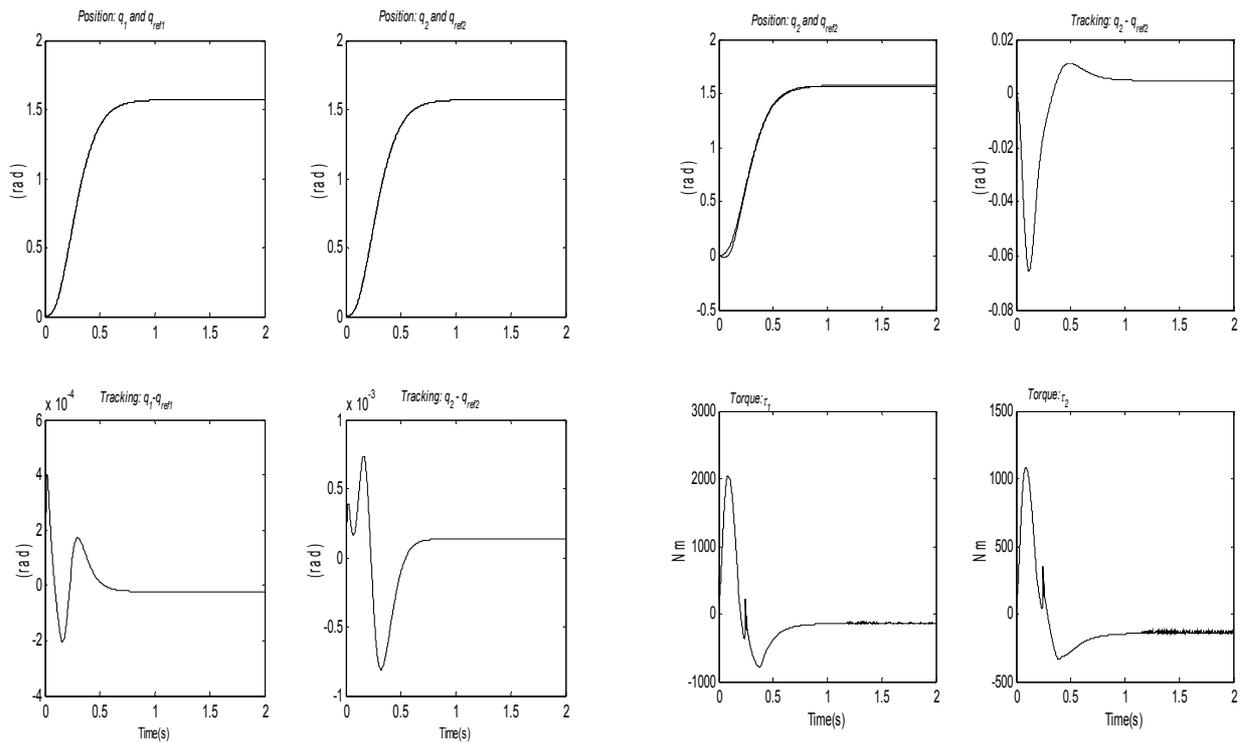

Fig.2 Position tracking error of joint $q_1$ and $q_2$. (matched case).

Fig.4 Position tracking performance in mismatched case.

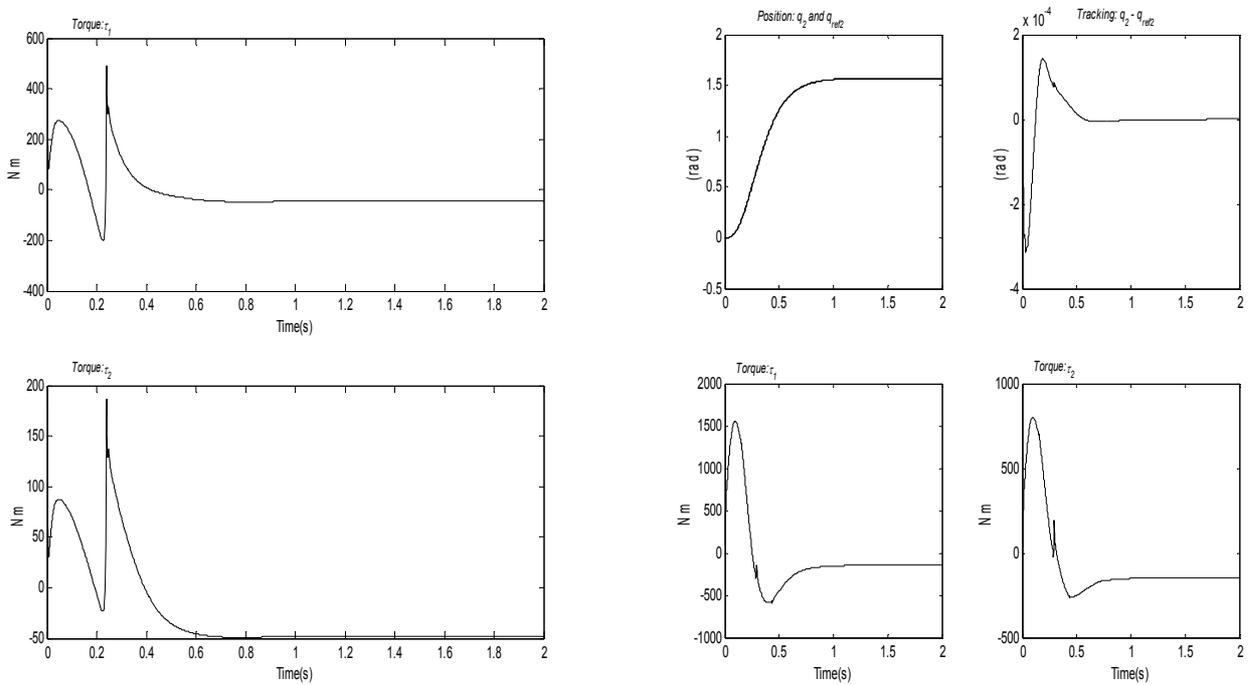

Fig.3 Induced torque control.

Fig.5 Torque and tracking performance with integral action.

In addition, other simulations have been carried out and the following remarks have been observed:

- Decreases in performances are obtained when we increase the step control parameter h .

- High dynamics of the reference trajectories results in high increase of the control torque signal. To reduce this control torque amplitude, one should increase the predictive time increment h. It should be pointed that over a threshold value of h noted $h_{max}$, the performance decrease and instability mechanism will appear. This is due to the Taylor approximation used to derive the predictive controller, which becomes invalid.

**7. Conclusion**

In this paper, the approximation of receding-horizon controller of rigid link robot manipulator using output feedback via link position measurements were considered. Minimizing a quadratic function of the predicted tracking error and the predicted input over the fixed horizon, by using the Simpson's rule approximation, derives the control law. One of the main advantages of these control schemes is that it does not need to perform an online optimization and asymptotic tracking of the smooth reference signal is guaranteed.

To enhance the robustness property of the nonlinear predictive developed by Ping Lu, we proposed to incorporate an integral action in the loop. Simulation shown that payload uncertainties and friction have no effect on the robot manipulator closed by the proposed algorithm. Moreover, the obtained torque signal lie in saturations limit. The Lyapunov theory is used to prove the asymptotic stability of equilibrium point of both original and augmented system.

Finally, we expect that the results presented here can be explored and extended to discrete implementation of these continuous-time predictive controllers either through computers or by special purpose chips that can run at a higher speed.